\documentclass[10pt,twocolumn,letterpaper]{article}

\usepackage{iccv}
\usepackage{times}
\usepackage{epsfig}
\usepackage{graphicx}
\usepackage{amsmath}
\usepackage{amssymb}
\usepackage{bm}

\usepackage[ruled,vlined]{algorithm2e}

\usepackage[caption=false]{subfig}

\usepackage[accsupp]{axessibility}  

\usepackage{array}
\newcolumntype{V}{>{\setbox0=\hbox\bgroup}c<{\egroup}@{\hspace*{-\tabcolsep}}}
\newcolumntype{C}[1]{>{\centering}m{#1}}

\usepackage[breaklinks=true,bookmarks=false]{hyperref}

\iccvfinalcopy 

\def\iccvPaperID{13} 

\ificcvfinal\fi

\begin{document}

\title{Encouraging Intra-Class Diversity Through a Reverse Contrastive Loss for Single-Source Domain Generalization}

\author{Thomas Duboudin$^1$,
Emmanuel Dellandréa$^1$,

Corentin Abgrall$^2$,

Gilles Hénaff$^2$,

Liming Chen$^1$\\

$^1$ LIRIS, École Centrale de Lyon, France\\
{\tt\small \{thomas.duboudin, emmanuel.dellandrea, liming.chen\}@ec-lyon.fr}\\
$^2$ Thales LAS France\\
{\tt\small \{corentin.abgrall, gilles.henaff\}@fr.thalesgroup.com}
}

\maketitle
\ificcvfinal\thispagestyle{empty}\fi

\begin{abstract}

Traditional deep learning algorithms often fail to generalize when they are tested outside of the domain of the training data. The issue can be mitigated by using unlabeled data from the target domain at training time, but because data distributions can change dynamically in real-life applications once a learned model is deployed, it is critical to create networks robust to unknown and unforeseen domain shifts. In this paper we focus on one of the reasons behind the inability of neural networks to be so: deep networks focus only on the most obvious, potentially spurious, clues to make their predictions and are blind to useful but slightly less efficient or more complex patterns. This behaviour has been identified and several methods partially addressed the issue. To investigate their effectiveness and limits, we first design a publicly available MNIST-based benchmark to precisely measure the ability of an algorithm to find the "hidden" patterns. Then, we evaluate state-of-the-art algorithms through our benchmark and show that the issue is largely unsolved. Finally, we propose a partially reversed contrastive loss to encourage intra-class diversity and find less strongly correlated patterns, whose efficiency is demonstrated by our experiments.

\end{abstract}

\section{Introduction}

While deep neural networks reach or even surpass human-level performance on an increasing number of computer vision tasks, \eg, classification, object detection or semantic segmentation \cite{he2016deep, he2017mask}, it is found that their performance could drop sharply \cite{hoffman2018cycada} when mismatch between training and testing data distributions occurs. This is an issue of critical importance for real-world application, where test-time domain shift is common: the data distribution can change over time because of varying factors as diverse as lighting, view angle, sensor, \etc, ending up being significantly different from the one used during training. For embedded networks deployed in the wild, that can't be easily retrained (autonomous cars, for instance), geographical distribution change are an issue too. It is impossible to gather enough data to cover all possible shifts, so another solution has to be found.\\

\begin{figure}[h]
\begin{center}
\includegraphics[width=1.0\linewidth]{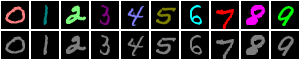}
\end{center}
   \caption{Samples of our benchmark. First row is training data, second row is testing data. For the training data, all digits are colorized with their class color, without exception. For the testing data, all digits have the same color, which is the average of the colors used in training.}
\label{fig:dataset}
\end{figure}

\begin{figure}[h]
\begin{center}
\includegraphics[width=0.75\linewidth]{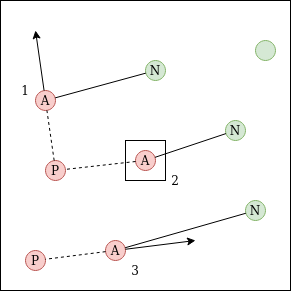}
\end{center}
   \caption{An illustration of our reverse contrastive approach on 3 anchors. Anchor 1 in the latent space is sufficiently far from its closest negative neighbor (full line) for it to be pushed away from its closest positive neighbor (dotted line). So is anchor 3. Anchor 2, however, is too close to a negative point to be pushed. The arrows show the moving direction of each anchor.}
\label{fig:principle}
\end{figure}

To generalize without having access to any kind of information regarding the target domain is the goal of the research field known as domain generalization (DG) (or \textit{out-of-distribution} generalization). Most of the works in this field assume to have access to data coming from several identified domains during training and aim to make the network's internal representations invariant on the domains, by finding patterns common to all domains. The idea behind is that patterns shared among several domains are more likely to be found in a new, unseen, domain \cite{li2018domain, moyer2018invariant}. This multi-source domain generalization (MDG) setting is not realistic for all tasks and all kind of data: health-related data, for instance, cannot be easily shared and the gathering of data from different sources (such as country) may prove difficult. In this paper, we are concerned with a situation, coined as single-source domain generalization (SDG), where only one source domain is available during training, either that all data come from a same distribution, or that the data come from several distributions which are not identified differently. Both settings (single and multi-source) are still largely unsolved in the general case: on several complex and realistic datasets, \eg, DomainNet \cite{peng2019moment}, PACS \cite{pacs}, methods designed to improve the out-of-distribution generalization capability of networks were found to perform as badly as the standard training procedure \cite{gulrajani2021in}.\\

The reasons behind the failure of deep networks to generalize outside of their training distribution are numerous and complex \cite{nagarajan2020understanding, shah2020pitfalls, hermann2020origins}. Among them is the inability of networks to use several semantically different clues, or patterns, to make their predictions and to instead rely only on the simplest most predictive patterns \cite{singla2021understanding, hermann2020shapes}. If the learned obvious class-specific patterns are spurious and therefore missing (or worse: anti-correlated or uncorrelated with the training class) at test time, the performance of the network will suffer. Furthermore, due to the nature of the data acquisition process, it is fairly common to observe such spurious correlations between patterns and labels in the data \cite{tommasi2015deeper, fabbrizzi2021survey}. Examples of such biases could be background and context, as in \cite{beery2018recognition}, or even inconspicuous hospital specific clues, as in \cite{degrave2021ai}. This behaviour, and its consequences on a network's ability to generalize outside of its training domain, have been previously identified and partially mitigated in \cite{huangRSC2020, pezeshki2020gradient}. However, using our benchmark specifically crafted to evaluate the ability of a model to find other patterns than the most strongly correlated ones, we show the inability of existing methods, \textit{e.g.}, Jigsaw \cite{carlucci2019domain}, Spectral Decoupling \cite{pezeshki2020gradient}, RSC \cite{huangRSC2020}, to correctly handle this issue. None of the existing works reach satisfying performance and some of them can record a performance drop as high as 56 accuracy points when the correlated pattern identified in training data, \textit{i.e.}, color, is missing in the test data. Note that we should not expect the discovery of "hidden" patterns to completely solve the SDG issue on real-life benchmarks, such as on PACS mentionned earlier: a pattern can be missing in the test-time distribution, but also uncorrelated or anti-correlated with its training class. Real-life situations are a complex blend of missing and uncorrelated (or anti-correlated) patterns. In such situations, it is not sufficient to learn the less strongly correlated patterns, some of them have to be ignored for the network to make a proper decision. Similarly, the performance of a DG method on a real-life benchmark is not an assessment of its ability to find the "hidden" patterns. Our benchmark, illustrated in Figure \ref{fig:dataset} is specifically crafted to avoid this issue.\\

Given these findings, we posit that, when training for a classification task, naturally learned patterns are learned because they tend to maximize inter-class specificity, while ignoring intra-class variability. These learned patterns are the most effective when it comes to discriminate different classes. As a result, we hypothesize that, to find less efficient patterns, we need to look for class-discriminant patterns with a higher intra-class variability than those learned naturally. That is, patterns enabling the network to better discriminate samples of the same class. Because we don't have access to additional auxiliary labels to discriminate elements within a class, the idea is to expand the class-wise internal representations of a deep network. We do so by using a partially reversed contrastive loss: in a traditional contrastive approach, same class samples are brought together in the latent space while different class samples are pushed away from each other. In our approach, same class samples are pushed away from each others until they reach a different class sample (see Figure \ref{fig:principle} for an illustration of our method). Our strategy yields state-of-the-art performance on our MNIST-based dataset.\\

Our contributions are threefold:

\begin{itemize}
\item We design an MNIST-based benchmark, named MNIST-MP (Missing Patterns), to study the ability of an algorithm to find other useful patterns than the naturally learned ones. It is available on github \footnote{\url{https://github.com/liris-tduboudin/Look-Beyond-Bias}}. 
\item We show that several state-of-the-art domain generalization algorithms especially designed to find "hidden" patterns fail to do so on a simple MNIST-based benchmark.

\item We propose a reverse contrastive loss (RCL) to find less strongly correlated patterns by encouraging intra-class diversity.
\end{itemize}

The remaining of the paper is organized as follows. Sec.2 discusses related works. Sec.3 introduces the proposed MNIST-based benchmark. Sec.4 presents our proposed RCL algorithm. Sec.5 benchmarks our proposed algorithm along with state-of-the-art algorithms for comparison. Sec.6 concludes the paper and draws some future works.

\section{Related Works}

\subsection{Domain Generalization (DG)}

\subsubsection{Robustness through multi source training}

Most settings in DG assume to have data coming from several different source domains, which are identified. By learning domain invariant features, through a great diversity of practical approaches, the model is supposedly able to generalize to a new domain. Most recent works use adversarial strategies: in the work of Tzeng \etal \cite{tzeng2017adversarial} the features extracted by a feature extractor are fed to a discriminator tasked to find the original domain of the image. The discriminator is trained to minimize the domain classification error, while the feature extractor is trained to maximize it, hence making the features indistinguishable between domains. This family of works were initially developed in the context of domain adaptation (DA) but can be adapted in the multi-source domain generalization setting by employing all the domains at disposal during training, \eg, \cite{carlucci2019hallucinating} and \cite{li2018domain}. The work of Krueger \etal \cite{krueger2020out} is enforcing invariance at loss level by minimizing the variance of the loss over all training domains alongside with the usual average of the loss per batch. Self-supervised strategies have also emerged to solve the problem of multi-source domain generalization. Carlucci \etal \cite{carlucci2019domain} used a self-supervised strategy based on solving jigsaw puzzles: the images are divided into tiles, which are shuffled before being given to the classifier. The classifier has two goals: classify the samples and find the shuffling permutation. By having a supplementary objective, not related in any way to the classification, the network will learn other patterns than those strictly sufficient to the classification task and hence generalize better to a new distribution. So far, it is unclear why features learned with self-supervised tasks are more robust to domain shifts. \\

In 2019, Gulrajani and Lopez-Par \cite{gulrajani2021in} analysed a lot of DG algorithms and benchmarks "in search of lost domain generalization". They detailed ways to carefully and realistically evaluate multi-source DG methods, by using larger models, stronger data augmentation, correct model and hyper-parameters selection (without any use of the target domain), and doing all on more datasets. Their conclusion was that no methods were significantly above the basic expected risk minimization procedure (ERM), where data from all domains is merged into one single dataset with the network trained on it. Their findings are corroborated by Cha \etal \cite{cha2021domain} who look into the latent spaces from differently trained networks, one with DANN \cite{ganin2015unsupervised} and another with the basic ERM. They have shown that ERM naturally tends to yield domain invariant features.

\subsubsection{Robustness through stylization}

In situations where only one domain is available during training (SDG), style transfer as a mean of data augmentation has been recently introduced independently by several authors \cite{nam2019reducing, yue2019domain, kim2020learning, somavarapu2020frustratingly} as a way to promote invariance to texture changes. The images are simply stylized differently, with AdaIN based \cite{huang2017arbitrary} or GAN-based \cite{NIPS2014_5ca3e9b1, zhu2017unpaired} architectures, before being fed to the main task model. It has been hypothesized that the low ability of deep networks to transfer to unseen domains was related to their texture bias: deep networks grant more importance to the texture than  the shape in image recognition tasks \cite{geirhos2018imagenet}. As such, correcting this texture bias is expected to improve out-of-distribution generalization, provided that the distribution shift is mainly due to a texture shift. The stylization as data augmentation exists in different flavors: Nam \etal \cite{nam2019reducing} use intra-domain extra-class stylization: the styles used to transfer are coming from the images of the training distribution but from different class samples than the image being stylized. Yue \etal \cite{yue2019domain} use several auxiliary style datasets. Jackson \etal \cite{jackson2019style} approximated the style codes distribution of paintings with a multivariate normal, and completely discard the painting dataset once the transfer network is trained. The stylization is a strong baseline in single-source domain generalization (SDG) \cite{yue2019domain} when the domain shift encountered is a texture shift, such as a synthetic to real transfer, which, unfortunately, does not always hold in the general case.

\subsubsection{Robustness by overcoming biases}

Deep neural networks learn correlations between patterns and labels no matter how spurious they seem to a human being and therefore sometimes need to be explicitly told not to use certain patterns for the recognition task at hand \cite{nam2020learning, Kim_2019_CVPR}. In a domain shift situation, the patterns naturally learned by a network are precisely the biases we need to overcome to find other and semantically different patterns. A line of work, \textit{e.g.}, Kim \etal \cite{Kim_2019_CVPR}, is able to create bias invariant features, but requires the bias to be given as an auxiliary label, and are therefore not suitable to the SDG setting where no information on testing data, \textit{e.g.}, bias shift, is assumed available. Another line of work, \textit{e.g.},  \cite{nam2020learning, dagaev2021toogoodtobetrue, ahmed2021systematic}, focuses on finding counter-examples \ie, samples that do not share the majority biases of their classes, and increase their importance during training. Assuming the necessary existence of counter-examples is an optimistic hypothesis in certain situations, such as with synthetic data where only a few number of synthetic models of interest may be available. Besides, it could lead to overstate the importance of what could be an aberrant outlier, such as an annotation error. As such, there also exists approaches which do not need counter-examples. Representation Self-Challenging \cite{huangRSC2020} (RSC) introduces a dropout-based strategy, where the feature map coefficients most responsible for the prediction are muted, instead of random ones, therefore leading the network to focus on other less correlated patterns. Spectral Decoupling \cite{pezeshki2020gradient}, a method introduced by Pezeshki \etal, proposes an L2 regularization on the output logits of the network, with a theoretical motivation, to push the network to learn other patterns than the obvious and spurious ones.

\subsection{Supervised Contrastive Learning}

Contrastive learning \cite{siamese2014taigman, hoffer2015deep} has been introduced in deep learning to enable class-wise manipulation of a network's latent space with a simple idea: elements of the same class should be close to each other in the latent space, elements of different classes should be away from each other in the latent space. Our approach is to loosen the latent space to identify less correlated patterns in a recognition task. While this idea has been previously studied in \cite{spread_out} and \cite{xuan2020hard}, it was applied with a different goal. Zhang \etal \cite{spread_out} proposes to take all the space available on the unit sphere evenly, but the regularization only impacts negative pairs. Xuan \etal \cite{xuan2020hard} aims to create a latent space with better generalization to unseen classes using a modified triplet loss that also pushes together elements of different classes. 

\section{MNIST "Missing-Patterns"}

\subsection{Training Set}

We introduce a new toy dataset based on the MNIST dataset \cite{mnist} to benchmark the ability of an algorithm to find less efficient but still useful patterns in the data. For the training data, each digit, coming from the original MNIST training data, is colorized with a color depending on its class. There are no counter-examples (digit colorized differently than the majority of the other digits in its class), as in \cite{ahmed2021systematic} or label noise (digit colorized with their correct class color, but with an assigned label different from the original one), as in \cite{pezeshki2020gradient}. We do so to study the ability of a network to find more semantically different patterns than what is done naturally, without having any kind of data signals that could help it find those patterns. The validation data is colorized the same way as the training data but come from the original MNIST test split. A sample of the dataset can be found Figure \ref{fig:dataset}. A summary of the synthetic MNIST-based DG datasets published so far is available in Table \ref{table:mnist}.

\begin{table}[h]
\begin{center}
\begin{tabular}{|l|c|c|c|c|}
\hline
Method & SDG/MDG & UP/MP & LN/CE & Classes \\
\hline\hline
IRM \cite{arjovsky2020invariant} &  MDG (2) & UP & LN & 2\\
\hline
SD \cite{pezeshki2020gradient} & SDG & UP & LN & 2 \\
\hline
GIP \cite{ahmed2021systematic} & SDG & UP \& MP & CE & 10\\
\hline
Ours & SDG & MP & None & 10 \\
\hline
\end{tabular}
\end{center}
\caption{Comparison of MNIST-datasets introduced to study domain generalization strategies. SDG/MDG refers to single or multiple training domains, UP/MP is uncorrelated (or anti-correlated) or missing patterns, LN is label noise, CE is counter-examples, Classes is the number of classes.}
\label{table:mnist}
\end{table}

\subsection{Testing Set}

The MNIST-MP dataset aims to benchmark SDG algorithms in presence of missing patterns, \textit{i.e.}, when a class correlated pattern in training, \textit{e.g.}, color, is missing in testing. It is made by coloring each digit with the same color, no matter their class. The images come from the original MNIST test data. The test color and the training colors are not chosen randomly. To precisely confront a deep network with a situation where useful training patterns are missing at test-time, the test color has to produce an activation that is roughly the same for all color specific filters in the network. This way, the network is unable to use color related information to predict the class of a digit and has to make use of other class correlated patterns, \textit{e.g.}, shape. Because we can't directly abide by this constraint we propose to use the average training color as the test color, provided that no specific training color is closer to the average than any other one and that all training colors are sufficiently different from each other. If the training color of class C is closer to the average color than the other colors, the network will wrongfully believe that test samples are all of class C. If two training colors are close to each other, the network won't be able to easily differentiate samples of the two classes using only the color and will be driven to learn other class related patterns, \textit{e.g.}, shape, at least partially. The colors (a triplet of value in range $[0,1]$) are therefore chosen to be on a sphere centered in the average (6 of them) and on the vertices of a cube centered in the average (the 4 left). This way no particular color is closer to the average than at least three others. All colors are not on the sphere to avoid colors that are too similar to one another.\\

\begin{table*}[h]
\begin{center}
\begin{tabular}{|l||c|C{3.1cm} VV|}
\hline
Method & Validation Accuracy &  Test Accuracy & Best Test - Val Acc & Best Test - Test Acc \\
\hline\hline
Standard Training Procedure & 99.8 ($\pm$ 0.006) & 26.0 ($\pm$ 4.7) & 99.7 ($\pm$ 0.08) & 31.9 ($\pm$ 5.5)\\
\hline
Dropout & 99.8 ($\pm$ 0.01) & 31.9 ($\pm$ 3.1) & 99.7 ($\pm$ 0.05) & 41.8 ($\pm$ 2.9) \\
\hline
Dropout \& Orthogonality \cite{ortho} & 99.8 ($\pm$ 0.008) & 42.7 ($\pm$ 2.6) & 99.7 ($\pm$ 0.03) & 44.1 ($\pm$ 2.5) \\
\hline
Dropout \& Covariance \cite{correlation} & 99.8 ($\pm$ 0.002) & 42.6 ($\pm$ 2.0) & 96.6 ($\pm$ 8.74) & 44.4 ($\pm$ 1.6) \\
\hline
Jigsaw Puzzle \cite{carlucci2019domain} & 99.8 ($\pm$ 0.01) & 43.0 ($\pm$ 3.0) & 98.6 ($\pm$ 0.6) & 58.6 ($\pm$ 4.7) \\
(with early stopping @95) & 98.5 ($\pm$ 0.3) & 59.9 ($\pm$ 4.5) & 98.5 ($\pm$ 0.34) & 59.0 ($\pm$ 4.35) \\
 \hline
Reconstruction & 99.8 ($\pm$ 0.007) & 28.5 ($\pm$ 4.8) & 99.7 ($\pm$ 0.1) & 31.6 ($\pm$ 5.0)  \\
\hline
Spectral Decoupling \cite{pezeshki2020gradient} & 99.8 ($\pm$ 0.004) & 47.7 ($\pm$ 2.9) & 99.7 ($\pm$ 0.06) & 51.0 ($\pm$ 1.5) \\
(with early stopping @95) & 99.4 ($\pm$ 0.09) & 49.8 ($\pm$ 1.5) & 99.4 ($\pm$ 0.08) & 48.9 ($\pm$ 2.2)\\
\hline
RSC \cite{huangRSC2020} & 99.2 ($\pm$ 0.2) & 43.5 ($\pm$ 5.0) & 97.6 ($\pm$ 2.8) & 48.0 ($\pm$ 3.6)\\
\hline
RCL-SDG (ours) &  &  &  &  \\
$m = 0.2$ & 99.8 ($\pm$ 0.007) & 12.5 ($\pm$ 2.6) & 99.8 ($\pm$ 0.01) & 15.3 ($\pm$ 5.4)\\
$m = 0.4$ & 99.8 ($\pm$ 0.007) & 19.9 ($\pm$ 5.5) & 99.8 ($\pm$ 0.1) & 27.1 ($\pm$ 6.4) \\
$m = 0.6$ & 99.8 ($\pm$ 0.009) & 36.8 ($\pm$ 5.6) & 99.4 ($\pm$ 0.4) & 49.8 ($\pm$ 4.7) \\
$m = 0.8$ & 99.7 ($\pm$ 0.03) & 55.7 ($\pm$ 4.7) & 98.2 ($\pm$ 0.7) & 73.2 ($\pm$ 3.0)\\
$m = 0.9$ & 99.4 ($\pm$ 0.08) &  68.2 ($\pm$ 4.0) & 97.6 ($\pm$ 1.12) & 76.2 ($\pm$ 6.8) \\
(with early stopping @95) & 95.84 ($\pm$ 0.6) & 74.7 ($\pm$ 8.5) & 95.9 ($\pm$ 0.7) & 76.9 ($\pm$ 2.0) \\
$\bm{m = \infty}$ & \textbf{96.0 ($\pm$ 0.3)} & \textbf{89.9 ($\pm$ 0.9)} & \textbf{96.0 ($\pm$ 0.6)} & \textbf{91.1 ($\pm$ 0.7)}\\ 
\hline
\hline
Standard Training Procedure & 97.8 ($\pm$ 0.12) & 97.8 ($\pm$ 0.12) & - & -\\
on the original MNIST & & & & \\
\hline
\end{tabular}
\end{center}

\caption{Results of the methods on the MNIST "Missing-Patterns" dataset. The results are obtained with a best validation selection strategy. The first column is the validation accuracy (colored digits), and the second the test (grey digits) accuracy. Accuracies reported are averages over 10 runs, with the standard deviation between parenthesis. The last line indicates the accuracy reached by our backbone on the original MNIST dataset without domain shift, it thus gives the highest bound of the accuracy in the test domain for the methods.}
\label{table:results}
\end{table*}

\section{Reverse Contrastive Approach}

\subsection{The proposed method}

\begin{figure*}[h]
\begin{center}
\setlength{\tabcolsep}{-7pt} 
\begin{tabular}{ccc}
\subfloat[]{\includegraphics[width=0.34\linewidth]{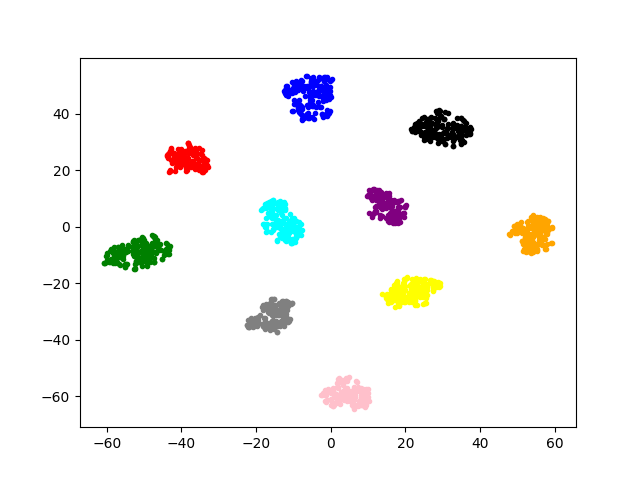}} &
\subfloat[]{\includegraphics[width=0.34\linewidth]{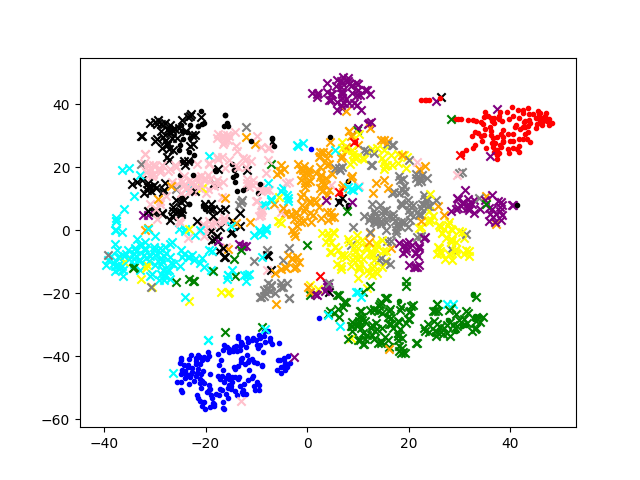}} & 
\subfloat[]{\includegraphics[width=0.34\linewidth]{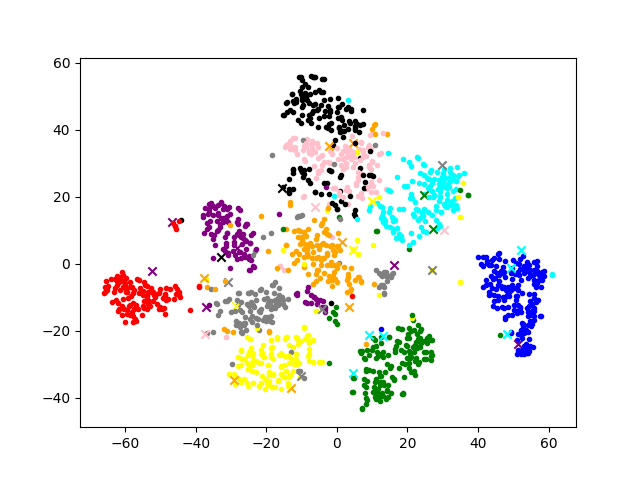}}\\[-0.4cm]
\subfloat[]{\includegraphics[width=0.34\linewidth]{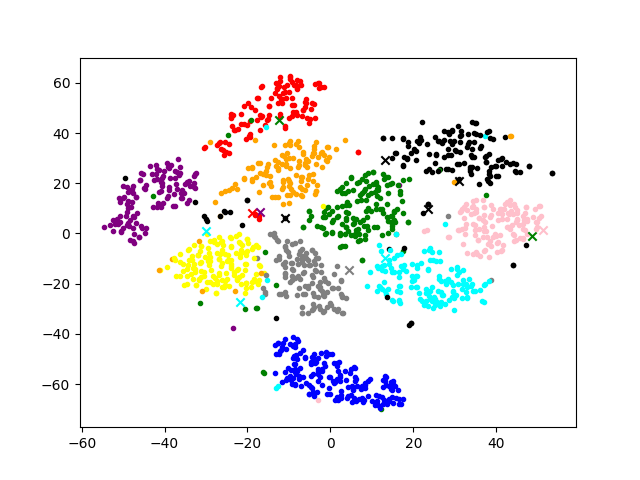}} &
\subfloat[]{\includegraphics[width=0.34\linewidth]{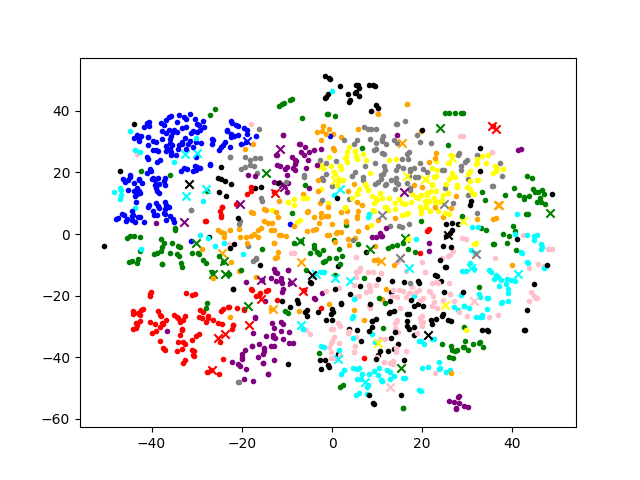}} &
\subfloat[]{\includegraphics[width=0.34\linewidth]{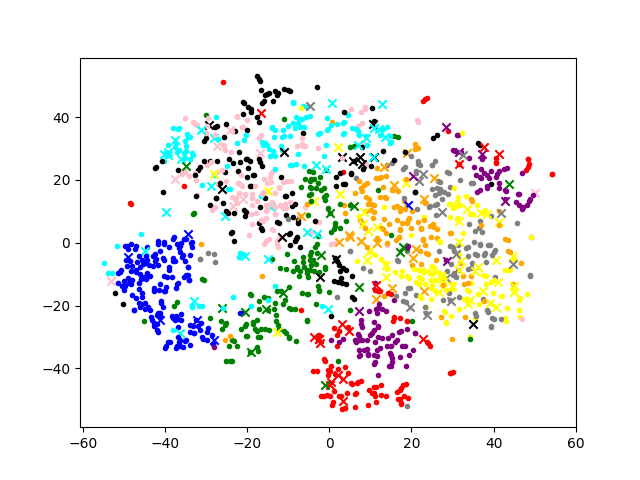}}\\
\end{tabular}
\vspace{3mm}
\caption{Visualisation of different latent spaces through a 2-dimensional T-SNE \cite{tsne}. The two axis are the dimensions obtained by the T-SNE. \textit{(a)} shows the validation (colored digits) latent space, for a model trained normally. \textit{(b)} shows the test (grey digits) latent space for the same model. \textit{(c)} shows the validation latent space for a model trained normally, on the original MNIST. \textit{(d)} shows the validation latent space for a model trained with the RCL and a margin of 0.9. \textit{(e)} shows the validation latent space for a model trained with the RCL without limitation. \textit{(f)} shows the test latent space for the same model. The color of a dot is its ground truth class (not the same as the colors used for the digits) and its shape represents whether or not the network successfully predicted the correct class: circle if so, cross if not. Best viewed in color.}
\label{fig:latent}
\end{center}
\end{figure*}

Our method starts from the observation that patterns that are learned with a standard training procedure (\eg, stochastic gradient descent with a cross-entropy loss, for a classification task) are learned because they exhibit the maximum inter-class specificity but they ignore intra-class diversity. To make the network learn semantically different ways to do the task, we propose to look for class-discriminant patterns with a higher intra-class variability than those learned naturally. A higher intra-class variability means that we are able to discriminate elements inside a single class, something that is difficult with the standard training procedure (see Figure \ref{fig:latent}, a). To find such patterns, we spread the intermediate features (the output of the convolutional features extractor F, before the fully connected classifier layers) of elements of the same class, to a certain extent. The limit in the class-wise repulsion is fixed with elements of the other classes: the intra-class features are repelled from each other until we reach elements of the other classes, with a margin. This approach is akin to a partially reversed constrastive loss where positive and negative pairs are switched, but a margin must be maintained between samples of different classes. We train the network with two objectives: the conventional classification objective for both the features extractor and the classifier, and the following for the features extractor only: 

\begin{equation}
    \begin{array}{ll}
    min_{F} \{ -d(f_a, f_p) \}\\
    \textit{ s.t. } d(f_a, f_p) < m \times d(f_a, f_n)
    \end{array}
\end{equation}

$f_a$ is the anchor feature map, $f_p$ a feature map belonging to a point of the same class as $f_a$ \ie the positive sample, $f_n$ a feature map belonging to a point of a different class than $f_a$, \ie the negative sample, and $m$ is the margin, a scalar to chose between 0 and 1 if we want the inter-class distance to be larger than the intra-class distance. We use a multiplicative margin, instead of an additive margin, contrary to the standard triplet loss \cite{hoffer2015deep}, to ease the hyper-parameter search. The distance $d$ used is the L1-distance. Feature maps are rescaled in the range $[0,1]$ before distance calculation, by using the maximum and minimum activation values computed batch-wise. This is done to avoid divergence as there are otherwise no lower bounds for this loss. Usually, features are normalized with their L2-norm, putting them on the unit sphere, but we found better results with the min/max strategy. Practically, we use the following reverse contrastive loss (RCL):

\begin{equation}
    \mathcal{L}_{RC} = \{
    \begin{array}{ll}
        -d(f_a, f_p) \textit{  if  } d(f_a, f_p) < m \times d(f_a, f_n) \\
        0 \textit{ otherwise}
    \end{array}
\end{equation}

The triplet of features used is not chosen randomly in the batch: we follow an easy-positive hard-negative sampling strategy. The anchors are sampled randomly, the positive is the closest element of the same class, the negative is also the closest element in a different class. It is useless to push away elements of the same class that are already far from each other in the latent space: patterns extracted for these samples are already different. We compare the closest intra-class distance with the closest inter-class distance to avoid creating a mixed latent space and define precise boundaries in the latent space. The positive points used in the loss are detached from the computation graph so that only the anchor is moved in the latent space: the distance check with negative neighbors is only relevant for the anchor. This loss cannot work on its own without the classification loss: at initialization, all features are clustered in the same area of the latent space and since the inter-class distances are small, we cannot expand the size of the intra-class clusters. The cross entropy loss will push away the features of elements of different classes, allowing the second objective to loosen the intra-class clusters. A visualisation of the proposed method can be found in Figure \ref{fig:principle} and the detailed algorithm in pseudo-code in Algorithm 1. We found even better results when pushing away features of the same class without any kind of limitation, that is, minimizing $-d(f_a, f_p)$, or having a margin set to infinity. While counter-intuitive, this might be explained by the influence of the classification loss, which prevents a complete scattering of intra-class features and forces the differentiation of inter-class features.

\begin{algorithm}
\small
\SetAlgoLined
method specific hyperparameters: \\
- weight for the RCL $\alpha$ \\
- margin for the RCL $m$\\
 \While{training is not over}{
  sample batch of data $\{(x_i,y_i), i=0...N$\}\\
  calculate cross-entropy loss on batch $\mathcal{L}_{CE}$\\
  calculate intermediate features  $\{f_i, i=0...N\}$\\
  normalize features\\
  \For{each sample $f_i$ in batch}{
  find closest positive $f_{p,i}$ in the batch\\
  find closest negative $f_{n,i}$ in the batch\\
  detach $f_{p,i}$ from the computation graph\\
  \eIf{$d(f_i, f_{p,i}) < m \times d(f_i, f_{n,i})$}
  {$\mathcal{L}_{RC,i} = -d(f_i, f_{p,i})$}
  {$\mathcal{L}_{RC,i}=0$}
  }
  $\mathcal{L}_{RC}=\frac{1}{N} \sum_{i=0...N} \mathcal{L}_{RC,i}$\\
  update model with $\mathcal{L}_{CE} + \alpha \times \mathcal{L}_{RC}$
  }
\caption{Reverse Contrastive Loss}
\end{algorithm}

\subsection{State-of-the-art baselines}

A first algorithm to find several useful patterns for classification can simply make use of dropout \cite{dropout}. By zeroing activations inside the network, we naively force it to look for new patterns. A limitation of dropout is that nothing prevents the network from learning the same patterns through several filters. To avoid this redundancy phenomenon, we further implement two  straightforward variants using two different regularizations: orthogonality of filters, used in \cite{ortho} (more precisely, we apply the double soft orthogonality regularization) and constraint over the covariance matrix of the filters activations, used in \cite{correlation}. Both regularizations are applied on the same layer as dropout, but before dropout is applied for the calculation. An orthogonality constraint for filters is supposed to prevent a filter to be close to another. The penalisation of the covariance matrix of the activations is based on the idea that filters that generally activate together, or don't activate together, are probably semantically related, even though their weights might be different. This constitutes what we call the naive strategies.\\

We also implement more elaborated methods inspired by DG works reviewed in sec.2. First, we compare our approach with one inspired from the jigsaw puzzle multi-task strategy used in \cite{carlucci2019domain}. Our approach is also compared to a reconstruction multi-task strategy (inspired from \cite{autoencoder}), where the features obtained by the features extractor are used for a classification task, with a classification head, and for an image reconstruction task, with a decoder. The feature extractor must extract sufficiently complete patterns to reconstruct the input, which are more than what is extracted with a single classification objective. The two strategies tailored precisely for the problem at hand are Representation Self Challenging \cite{huangRSC2020}, and Spectral Decoupling \cite{pezeshki2020gradient}. The selected methods were chosen because they improve out-of-domain generalization by finding new patterns. We did not compare ourselves with more general methods, such as these using style transfer for instance, because our goal is to measure the ability of an algorithm to find less correlated patterns.

\section{Results and Discussion}

\subsection{Experimental setup}

Our experiments are conducted on our benchmark using a small neural network. The architecture we use was introduced in \cite{carlucci2019domain} to study MNIST to SVHN \cite{svhn} transfer. It is composed of two convolutional layers and three fully connected layers. Max pooling operations are inserted between each convolutional. The non-linearity used is ReLU for all the layers. The convolutional layers define the feature extractor (128 channels), and the fully connected layers the classifier. For all experiments, we use stochastic gradient descent, with a batch size of 128, with nesterov momentum at 0.9, a fixed learning rate of 1e-3 and an L2 weight decay at 1e-5. Models are trained for 10 epochs. There is no data augmentation. The jigsaw puzzle strategy uses an additional fully connected layer to the network, as in \cite{carlucci2019domain}, alongside another fully connected layer used for the classification. The images are divided into 2x2 squared tiles, which are then shuffled, yielding 24 possible permutations. Each batch is used for both labels: class with the original batch sent through the network, permutation with the shuffled batch. The decoder in the reconstruction method is composed of 4 transposed convolution layers, with an hyperbolic tangent as the last activation function. When dropout is used, the dropout rate is chosen at random between 0 and 1 for each iteration: a fixed dropout rate helps the network introduce redundancy in a simple fashion: it only has to create more redundancies than there are dropped channels. Likewise, only full channel dropout is used because of the correlation between spatially close activations in the same channel, which enable the network to recover the color all the time. For RSC, we reuse these dropout hyper-parameters and the batch percentage (the proportion of samples per patch for which RSC is used) is fixed at 100\% as we want the network to look beyond the color for every image. \\

Most strategies use two objectives during training: the classification cross-entropy and another objective (jigsaw puzzle, reverse contrastive loss, reconstruction) or regularization (orthogonality, spectral decoupling). The weight for the classification loss is always set to 1. The weightings for the supplementary losses (or regularizations) were selected in the list $(0.001, 0.01, 0.1, 1, 5, 10)$ with the following principle: the value selected is the largest value that does not lead to a collapse of the validation accuracy. The idea is that the regularization weighting should have a positive slope for out-of-distribution accuracy, \ie the larger the weight, the better the out-of-distribution accuracy, up until a certain point. It is grounded in the fact that most additional objectives tend to counter the natural behaviour of the network. The weight is 1.0 for the orthogonality constraint, 0.01 for the covariance constraint, 10.0 for the Jigsaw Puzzle, 1.0 for the reconstruction, 5.0 for Spectral Decoupling and 1.0 for our reverse contrastive constraint. \\

During training, we select the model with the highest validation accuracy, and evaluate this model on the testing data. It has been noted that domain generalization is a setting where the initialization of the network is more important than usual and that results may vary in a greater fashion than with a training and testing dataset coming from the same distribution \cite{krueger2020out}. Therefore, we average the results over 10 runs, and report the standard deviation alongside the average.

\subsection{Results and analysis}

Table \ref{table:results} synthesizes the overall results. As can be seen, our reverse contrastive method yields a significant improvement over the previous works and the naive strategies on our dataset. Dropout compares favorably to normal training, but yields far better results when used together with a regularization to prevent redundancy. This redundancy issue might explain why RSC yields results only marginally above dropout and regularization. During training, we monitored the test accuracy over the epochs and noticed that jigsaw puzzle and spectral decoupling succeed to some extent but suffers from an over-fitting issue: the best test accuracy does not happen for the model with the best validation accuracy. Early stopping is useful in this situation. We employ a simple yet realistic early stopping strategy that does not use the test set: training is stopped as soon as the validation accuracy reaches a satisfying threshold, fixed here at 95\%. This only enables us to recover a closer accuracy than the best test accuracy but not go higher. Most methods specifically designed to prevent the network to focus only on a subset of useful features are not effective enough to consider the problem solved on our simple dataset. \\

Beside the accuracy table, we also illustrate the impact of our reverse contrastive loss in Figure \ref{fig:latent} by looking at the latent space extracted from models trained differently, through the T-SNE dimensionality reduction method \cite{tsne} applied on the features. With a margin of 0.9, the class clusters are still separated but we can see their expansions (d), compared to the standard training situation (a). Without limitation, the clusters are mixed together, and can only be discriminated one another only roughly (e). This explains why the performance in the training domain is lower: the classifier can't perfectly separate the classes, however the performance in the testing domain is higher. This illustrates an underlying trade-off in our method: the more the intra-class clusters are expanded, the more the network will find "hidden" useful patterns but the more it will also learn useless patterns, that are not inter-class discriminant. By comparing with the latent space of the same model trained on the original MNIST (c), we see an inefficiency of the method to find useful patterns. The cluster's sizes needed to obtain an accuracy around 90\% are larger than the size of the clusters on the original MNIST, indicating that noise and instance specific patterns have been learned by the network. We hypothesize that on more complex datasets, the RCL without limitation might not yield good results due to the images specificities being more prevalent than in MNIST. To tackle this issue is a future work direction.\\

Our approach tends to qualify the common principle in the deep learning research community that a good latent space, one able to generalize well, is supposed to have tight intra-class clusters with large margins between clusters. While this is true when there is no domain shift, it might not always hold true when so. This can be seen in Figure \ref{fig:latent} where a testing latent space corresponding to a large-margin tight-clusters training latent space is completely misunderstood by the network (b). On the opposite, the blurred boundaries training latent space (e) stays similar in the testing domain (f).

\section{Conclusion}

In this paper we carefully created a benchmark to study one of the reasons deep networks fail to generalize outside of their training domain: the reliance of a model on only the most obvious discriminant patterns has dramatic consequences if, in the test domain, such patterns are missing. We showed that existing methods only mitigate the damage. Therefore, we proposed a counter-intuitive strategy: instead of aiming for a tight-cluster large-margin latent space, it is beneficial to try to expand the class-wise clusters, as the cluster-size is linked to the diversity of patterns learned. Experiments have shown that our approach performs significantly better than state-of-the-art approaches on our benchmark. The goal was not to compete directly against other SDG methods on realistic benchmarks, but to provide a building block for a future general SDG algorithm. The future works will be dedicated to the study of more real-life-like situations, and to the decorrelation (or anti-correlation) issue. Our benchmark MNIST-MP is publicly available on github, together with an MNIST-UP also proposed for the anti-correlation studies.

\section*{Acknowledgment}

This work was in part supported by the 4D Vision project funded by the Partner University Fund (PUF), a FACE program, as well as  Alegoria project and Arès labcom projects, both funded by the French Research Agency, l’Agence Nationale de Recherche (ANR), under grants ANR-17-CE38-0014-03 and ANR-16-LCV2-0012-01, respectively.

\newpage
{\small
\bibliographystyle{ieee_fullname}
\bibliography{egbib}
}

\end{document}